\documentclass[authorversion,nonacm]{acmart}
\setcopyright{cc}
\setcctype{by}
\usepackage{multirow}
\usepackage{svg}
\usepackage{array}

\AtBeginDocument{%
  }

\begin{document}

\title[Improving Patient Subtyping]{Improving Patient Subtyping on Longitudinal Data using Representations from Mamba-based Architecture}

\author{Md Mozaharul Mottalib}
  \affiliation{
  \institution{University of Delaware}
  \city{Newark}
  \state{Delaware}
  \country{USA}
}
\email{mmmdip@udel.edu}
\orcid{0000-0003-4930-0365}

\author{Rahmatollah Beheshti}
\affiliation{
  \institution{University of Delaware}
  \city{Newark}
  \state{Delaware}
  \country{USA}
}
\email{rbi@udel.edu}
\orcid{0000-0001-8912-3063}

\renewcommand{\shortauthors}{Mottalib et al.}

\begin{abstract}
Effective sub-typing (also known as grouping or clustering) of patients using their electronic health record (EHR) data can greatly inform precision medicine efforts. However, subtyping temporal EHR datasets is known to be challenging due to inherent EHR issues, including complexity and irregularity. In this study, we propose a self-supervised Mamba-based model that learns effective EHR representations and enables enhanced patient subtyping. We evaluate the proposed model on public and private real-world EHR datasets to classify the data based on the available labels and subtype patients based on the representations learned from the model. Through an extensive set of experiments, we demonstrate that our model's design choices lead to better performance compared to competitive baseline models for prediction. Moreover, we evaluate several clustering techniques to demonstrate that our findings offer valuable insights into subtyping patients based on temporal records from EHR models\footnote{Our implementations are available at \url{https://github.com/healthylaife/triplet_mamba}.}.
\end{abstract}



\maketitle

\section{Introduction}
Physicians rely on comprehensive medical histories, including laboratory results, procedures, medications, and diagnoses, to determine appropriate therapeutic interventions. Beyond individual clinical decision making, researchers utilize these longitudinal records and demographic data to identify cohort patterns, evaluate medication efficacy, and analyze disease prognosis. A critical methodology in this domain is patient subtyping, which seeks to identify groups of patients with shared disease progression pathways to address inherent heterogeneity and facilitate precision medicine. By identifying shared characteristics within complex longitudinal trajectories, these models can reveal hidden patterns of disease \citep{rajkomar2018scalable, li2020behrt, mottalib2023subtyping}, significantly enhancing the personalization of treatment plans and the prediction of early disease onset \citep{lee2020temporal, fredriksen2024categorization}. Effectively categorizing these trajectories is not merely a predictive exercise but a functional requirement for improving clinical outcomes through targeted intervention.

Notwithstanding the critical importance of the task, subtyping temporal datasets is exceptionally challenging due to the inherent nature of healthcare data, which is characterized by missingness, extreme sparsity, and irregular sampling. While transformer-based models have seen success in clinical prediction \citep{pang2021cehr, li2020behrt, rupp2023exbehrt, rasmy2021med, li2022hi, labach2023duett, poulain2024graph}, their quadratic computational complexity ($O(L^2)$) creates a critical bottleneck \citep{vaswani2017attention}. This scaling property effectively limits the context length, the amount of history the model can ingest, forcing a choice between truncating valuable patient records or adopting computationally prohibitive discretization methods like binning or imputation \citep{shuklainterpolation, che2018recurrent}. Such limitations are particularly detrimental to subtyping, which relies on capturing long-range healthcare dynamics to identify subtle transitions in a patient's state.

To address these challenges, we propose \texttt{Triplet-Mamba}, a self-supervised architecture designed for longitudinal representation learning and enhanced patient subtyping. Our model transitions from the traditional transformer to a Mamba-based backbone, utilizing a selective State Space Model (SSM) to achieve linear-time complexity ($O(L)$) \citep{gumamba}. This shift is a functional necessity for processing the long patient trajectories required to define robust clinical phenotypes without exceeding memory constraints \citep{gumamba}. The SSM backbone acts as an efficient continuous-time filter, which is uniquely suited for the non-uniform sampling rates inherent in EHR data.

Unlike prior methods that discretize data, \texttt{Triplet-Mamba} handles irregular sampling natively by embedding continuous timestamps and measured values directly. By training these representations with a self-supervised forecasting objective, the model learns a structured latent space that captures genuine disease progression rather than simple predictive memorization. We demonstrate through extensive experiments on public \citep{silva2012predicting, johnson2020mimic} and private datasets that this design yields more coherent and clinically relevant patient subtypes than competitive baseline models \citep{tipirneni2022self, labach2023duett}.

The main contributions of our work can be summarized as follows:
\begin{itemize}
    \item A novel Mamba-based architecture, \texttt{Triplet-Mamba}, which addresses the requirement of extensive context lengths to effectively capture longitudinal healthcare dynamics.
    \item The design of an input representation for this architecture that embeds continuous timestamps and measured values, preserving fine-grained information.
    \item A novel patient subtyping pipeline that learns patient time series representation through a self-supervised forecasting task and categorizes time series into meaningful subtypes.
\end{itemize}

\subsection*{Generalizable Insights about Machine Learning in the Context of Healthcare}


This study illustrates that effective and efficient patient subtyping in healthcare depends critically on learning scalable, temporally expressive representations that align with the structure of longitudinal EHR data. By prioritizing linear-time sequence models and self-supervised objectives over task-specific supervision alone, the study demonstrates that long, irregular patient trajectories can be modeled without sacrificing computational feasibility or clinical fidelity. A central insight is that robust representation learning that is supported by principled event-level encodings and self-supervised forecasting yields latent spaces that enable more coherent, stable, and clinically meaningful patient subtypes across datasets and clustering methods. More broadly, the work emphasizes that advances in patient stratification arise from jointly considering efficiency, temporal structure, and evaluation focused on subtype quality, reinforcing subtyping as a foundational, representation-driven problem in machine learning for healthcare.

\section{Related Work}

\paragraph{Transformers to State Space Models}
The success of the transformer architecture in natural language processing spurred its adoption for clinical data, and models like STraTS \citep{tipirneni2022self} and DuETT \citep{labach2023duett} made significant strides by designing specialized input representations and attention mechanisms. 
Despite their power, a critical bottleneck of Transformer-based models is the quadratic computational complexity \citep{vaswani2017attention} of the self-attention mechanism, which limits the length of patient histories that can be efficiently processed \citep{rajkomar2018scalable, wen2023transformers, li2022hi}. This limitation could be significant in healthcare, as a longer context often improves clinical prediction performance \citep{wornowcontext}.

This challenge has motivated the development of sub-quadratic architectures, most notably State Space Models (SSMs) \citep{gu2020hippo, fu2023simple, poli2023hyena, dao2023hungry}. The recent introduction of Mamba \citep{gumamba}, a selective SSM, has achieved performance competitive with Transformers but with linear-time complexity, making it highly efficient for very long sequences. Mamba's potential in healthcare is actively being explored \citep{mottalib2025hymate}. For instance, EHRMamba \citep{fallahpour2024ehrmamba} adapts the architecture for EHR forecasting tasks, and ClinicalMamba \citep{yang2024clinicalmamba} has been developed as a foundation model pre-trained on large corpora of clinical notes. Our work builds on this momentum, leveraging Mamba's efficiency for learning from long patient trajectories.

\paragraph{Learning Representations for Patient Subtyping}
A primary clinical application for these powerful representations is patient subtyping—the unsupervised grouping of patients into clinically meaningful phenotypes based on their disease progression. Effective subtyping can inform precision medicine and reveal novel insights into complex diseases. Recent computational approaches \citep{gorla2025epigenetic, jiang2023age} have proven instrumental in elucidating and validating novel phenotypes, enabling patient stratification based on both static, cross-sectional snapshots and dynamic, longitudinal trajectories.

A variety of deep learning techniques have been applied to this problem. Early works used Recurrent Neural Networks, such as the Time-Aware LSTM (T-LSTM) \citep{baytas2017patient}, to learn patient representations for clustering. Other approaches have utilized autoencoders \citep{ienco2020deep}, sometimes enhanced with attention mechanisms \citep{li2025autoencoder, li2023multimodal}, to learn similarity measures for subtyping from time-series information. More recently, research has also explored multi-modal data; for example, M-ClustEHR \citep{bampa2024m} employs a multi-modal autoencoder to learn comprehensive representations from diverse EHR data for sepsis subtyping. X-DEC model validated a Deep Embedded Clustering model and its adaptation for integrating mixed datatypes with an X-shaped variational autoencoder \citep{de2024deep}. Med-ROAR \citep{baskett2025identifying} hierarchically partitions data by encoding it into multi-level discrete representations, employing a modified self-attention mechanism for identifying homogeneous patients.

Our model, \texttt{Triplet-Mamba}, synthesizes these threads of research. We utilize the architectural efficiency of Mamba to effectively model long and irregular patient histories, combined with a specialized triplet input representation designed to capture fine-grained temporal details. By training these powerful representations with a self-supervised objective, we aim to produce more coherent and clinically relevant patient subtypes than those achieved by prior methods.

\section{Methods}

\paragraph{Problem Definition}
Consider a dataset structure that corresponds to typical EHR records for patients. For each patient, the record contains a time series of events corresponding to irregular patient observations, such as vitals and lab results, and a set of static variables that do not change over time, such as demographics. This can be represented as a sparse irregular time series dataset of the form $\mathcal{D} = {\{(\mathcal{s}^k, \mathcal{T}^k, y^k)\}}_{k=1}^N$ consisting of $N$ labeled samples, where the $k$-th sample contains a static vector $\mathcal{s}^k \in \mathbb{R}^{n_s}$ of $n_s$ static variables, a multivariate time-series $\mathcal{T}^k=(t_1^k, t_2^k,\cdots,t_{n_k}^k)$ of variable length $n_k$ and a corresponding binary label $y^k \in \{0,1\}$ for predefined task, e.g. mortality or condition prevalence.

Each event $t_i^k$ is an observation triplet, formally defined as a triplet $(t, f, v)$, where $t \in \mathbb{R}_{\geq 0}$ represents the time, $f \in \mathcal{F}$ denotes the feature or variable, and $v \in \mathbb{R}$ signifies the value of the observation, for example, \texttt{[72.9 days, body\_weight, 153lbs]}.  Omitting $k$ for the $k$-th patient, a multivariate time-series $\mathcal{T}$ of length $n$ is thus defined as a set of $n$ observation triplets, i.e., $\mathcal{T} = \{{(t_i , f_i ,v_i)}\}_{i=1}^{n}$. The underlying set of time-series variables, denoted by $\mathcal{F}$, may include vital signs (such as temperature), lab measurements (such as hemoglobin levels, body weight, etc.), and input/output events (such as fluid intake and urine output).  Thus, for a patient $p$ the target aims at predicting $y^p$ given $(\mathcal{s}^p, \mathcal{T}^p)$, the static variables $\mathcal{s}^p$ and multivariate time series $\mathcal{T}^p$.

\begin{figure}[htbp]
    \centering 
    \includegraphics[width=0.6\textwidth]{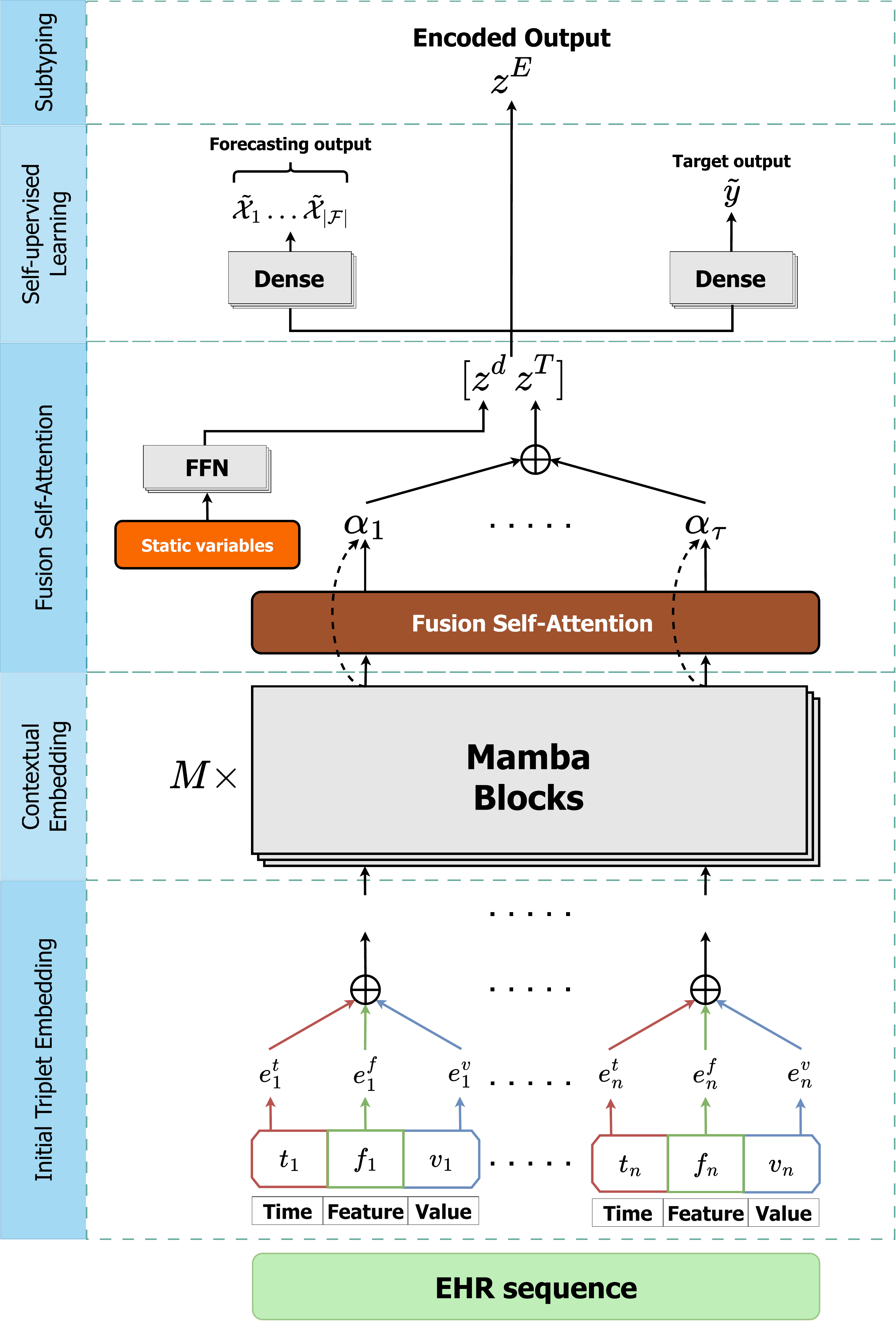}
    \caption{\texttt{Triplet-Mamba} architecture}
    \label{fig:model_arch} 
\end{figure}

\subsection{Model Architecture}
The architecture of our model, \texttt{Triplet-Mamba}, is illustrated in Figure \ref{fig:model_arch}. 
We define the input as a set of observation triplets, where each observation triplet in the input is embedded using an \textit{Initial Triplet Embedding module}. The initial triplet embeddings are then passed through a \textit{Contextual Triplet Embedding} module, which utilizes the Mamba blocks to encode the context for each triplet. The \textit{Fusion Self-attention} module then combines these contextual embeddings via a self-attention mechanism to generate an embedding for the input time-series, which is concatenated with a static (demographics) embedding and passed through a prediction layer to make the final prediction. We introduce subtyping as a downstream task from learned representations.

\subsubsection{Initial Triplet Embedding} An \textit{Initial Triplet Embedding} maps each of the $n$ triplets of the input sequence $\mathcal{T}={\{(t_i,f_i, v_i)\}}_{i=1}^n$ into a $\tau$-dimensional vector $e_i \in \mathbb{R}^\tau$, where $\tau$ is a hyperparameter for the dimension of the mapped variables. The vector $e_i$ is calculated by summing all embeddings into a single embedding, such that $e_i = e_i^f + e_i^v + e_i^t\in \mathbb{R}^{\tau}$. 

The embedding for the variable, $e_i^f$, is found using a look-up table mechanism similar to word embeddings. For the continuous value and time embeddings, $e_i^v$ and $e_i^t$, one-to-many Feed Forward Networks (FFN) with learnable parameters are used to calculate each embedding: $e_i^v = FFN^v(v_i)$ and $e_i^t = FFN^t(t_i)$. Both FFNs have a single input and a $\tau$-dimension output, with one hidden layer containing $\left\lfloor\sqrt{\tau}\right\rfloor$ nodes and a $tanh(\cdot)$ activation function. Unlike sinusoidal encoding with fixed frequencies \citep{yin2020identifying}, this approach enhances flexibility by enabling end-to-end learning of continuous value and time embeddings. 

While in EHRMamba \citep{fallahpour2024ehrmamba} event tokens are represented with seven embeddings with certain embedding repetitions due to their type, such as age embedding, we adapted the method of representing events in terms of triplets \citep{horn2020set, tipirneni2022self, wornowcontext}. This allows for adaptive representations that capture the complexities of the data, thereby improving performance in patient clustering and forecasting tasks. 

\subsubsection{Contextual Triplet Embedding} The initial triplet embeddings $\{e_1, e_2, ... ,e_n\}$ are then passed to the \textit{Contextual Triplet Embedding} module, which consists of $M$ blocks with $h$ Mamba layers. Each block takes $n$ input embeddings $\mathcal{E}\in \mathbb{R}^{n\times \tau}$ and applies normalization using root mean square normalization (RMS normalization). Then, the tensor is expanded through two linear projections, one of which undergoes a convolution followed by a Sigmoid Linear Unit (SiLU) activation, and its output is processed by the discretized state space model (SSM) to filter relevant information. 

The other projection is directly passed through a SiLU activation and combined with the outputs from the SSM using a multiplicative gate. This combined output is then passed through another linear projection and summed with the initial input $\mathcal{E}$, resulting in the final output embeddings $\mathcal{C}_{emb}\in \mathbb{R}^{n\times \tau}$. For additional details about the SSM unit, we refer to Appendix \ref{apd:mamba_block}.

\subsubsection{Fusion Self-Attention} From the last Mamba block, the output, $\mathcal{C}_{emb}=\{c_1, c_2, \dots, c_n\}$, is passed to the \textit{Fusion Self-attention}. Here, an FFN is applied to every contextual embedding $c_i$, followed by a softmax to calculate attention weights, $\alpha_i$. The final embedding for the time-series can then be computed by a weighted sum of the contextual triplet embeddings and the attention weights, using the following process, 

\begin{equation}
    a_i=\textbf{u}_a^T \tanh{(\textbf{W}_a \textbf{c}_i +\textbf{b}_a)}
\end{equation}
\begin{equation}
    \alpha_i={\frac{\exp{(a_i)}}{\sum_{j=1}^\tau \exp{(a_j)}} \hspace{2em}} \forall i=1,\dots,\tau
\end{equation}
\begin{equation}
    z^T=\sum^\tau_{i=1} \alpha_i \textbf{c}_i,
\end{equation}
where, $\textbf{W}_a\in \mathbb{R}^{d_1\times \tau}, \textbf{b}_a\in \mathbb{R}^{d_a},\textbf{u}_a\in \mathbb{R}^{d_a}$ are the weights of this attention network, which has $d_a$ neurons in the hidden layer. 
For the static (i.e., demographic) variables, the embedding, $z^d$, is generated using an FFN with 2 hidden layers, each containing $(2 \times d)$ nodes, where $d$ is the hyperparameter controlling the size of the embedded demographic variable vector. We then encode variable-length time-series to a fixed-length vector, $z^E $, concatenating demographic variable embeddings $z^d$ and time variable embeddings $z^T$, 
\begin{equation}
    z^E = \begin{bmatrix}
        z^d \\ z^T
    \end{bmatrix} \in \mathbb{R}^{d_{emb}}
\end{equation}where $d_{emb} = \tau + d$, represents the sum of dimensions for the time embeddings $\tau$ and demographic embeddings $d$.

\subsection{Training}
The model is trained in two phases, described below.

\subsubsection{Self-supervised pre-training} Our model incorporates forecasting as a self-supervision task using a bigger dataset with $N'\ge N$ samples given by $\mathcal{D}'={\{(\textbf{s}^k, \textbf{T}^k, \textbf{m}^k, \mathcal{X}^k)\}}_{k=1}^{N'}$. Here, $\textbf{m}^k \in {\{0,1\}}^{|\mathcal{F}|}$ is the forecast mask which indicates whether each variable was observed in the forecast window and $\mathcal{X}^k    \in \mathbb{R}^{|\mathcal{F}|}$ contains the corresponding variable values when observed. Also, $\mathcal{F}$ is the underlying set of time-series variables. The forecast mask plays a crucial role by enabling the exclusion of unobserved forecasts during the training process, ensuring that only relevant data informs the loss function. The forecasting task uses the learned patient representation $\tilde{\mathcal{X}}$,
\begin{equation}
    \tilde{\mathcal{X}} = w_s\begin{bmatrix}
        z^d \\ z^T
    \end{bmatrix}+b_s \in \mathbb{R}^{|\mathcal{F}|},
\end{equation}obtained from passing the concatenated embeddings through a dense layer with weights $w_s   \in \mathbb{R}^{|\mathcal{F}|}, b_s  \in \mathbb{R}$. The masked MSE loss is used for training on the forecasting task to account for missing values in the forecast outputs. Thus, the loss for self-supervised forecasting can be given by, 
\begin{equation}
    \mathcal{L}_{ss}=\frac{1}{|N'|}\sum^{N'}_{k=1}\sum^{|\mathcal{F}|}_{j=1}{m_j^k{(\tilde{\mathcal{X}}_j^k-\mathcal{X}^k_j)}^2},
\end{equation}where $m_j^k \in \{0,1\}$, based on the ground truth forecast $\mathcal{X}^j_k$ unavailability or availability respectively for $j$-th variable in $k$-th sample. 

\subsubsection{Fine-tuning} During fine-tuning, we use the patient representation $z^E$ extracted from the fusion self-attention and attach heads tailored to the downstream tasks. The final prediction for the target task is obtained by passing the concatenated embeddings through a dense layer with weights $w_o^T\in \mathbb{R}^{d_{emb}}, b_o\in \mathbb{R}$ and a sigmoid activation:\begin{equation}
    \tilde{y}=sigmoid(w_o^T\begin{bmatrix}
        z^d \\ z^T
    \end{bmatrix}+b_o).
\end{equation}The model is trained on the target using cross-entropy loss for the target task of binary classification.

\section{Experiments} 
In this section, we present a comparative analysis of the performance of our model and illustrate its effectiveness for subtyping. 

\subsection{Cohort}
We evaluated our proposed model on three EHR datasets. In particular, we used two widely used EHR datasets: PhysioNet challenge \citep{silva2012predicting} and MIMIC-IV \citep{johnson2020mimic, pmlr-v193-gupta22a, adiba2026multimodaldataprocessingpipeline}. PhysioNet challenge is a standardized public dataset with the task of predicting in-hospital mortality after the first $48$ hours of patient stays in the ICU, with $14.2\%$ of positive labels. For MIMIC-IV, we evaluated tasks on a derived ICU dataset following prior work by \citet{labach2023duett}, containing $53,150$ patients with $69,211$ admissions. Furthermore, we evaluated our model on a private pediatric weight management dataset extracted from the EHR of a large pediatric healthcare system providing primary, specialty, inpatient, and emergency care to pediatric patients across five US states. The detailed descriptions about the dataset can be found in Appendix \ref{apd:extended_cohort}. 

\subsection{Baseline Methods}
Among the nine baseline models evaluated, we discuss the top three performers, distinguished by their superior performance metrics. A comprehensive table detailing the results of all the baseline models, along with their complete performance statistics, is present in Appendix \ref{apd:baseline_model_comp}.
\subsubsection{\texttt{STraTS}: The Self-supervised Transformer for Time-Series \citep{tipirneni2022self}} uses a continuous value embedding technique to encode continuous time and variable values without discretization, featuring a Transformer component with multi-head attention layers.
\subsubsection{\texttt{EHR-Mamba} \citep{fallahpour2024ehrmamba}} encodes EHR data in a combination of seven different embeddings and uses stacked Mamba blocks for mapping input sequence to output tensor for downstream forecasting or predicting tasks.
\subsubsection{\texttt{DuETT}: The Dual Event Time Transformer for Electronic Health Records \citep{labach2023duett}} extends transformers to exploit both time and event modalities of EHR data. Each DuETT layer consists of two transformer sub-layers that attend and process along the event and time dimensions, respectively.

\subsection{Evaluation Metrics}
Model performance for the binary predictive tasks was evaluated using the Area Under the Receiver Operating Characteristic Curve (AUROC) and the Area Under the Precision-Recall Curve (AUPRC). Experiments were independently conducted $10$ times with distinct randomized seeds. The reported performance metrics represent the average and standard deviation across these repetitions. Statistical significance of the results was assessed using independent two-sample t-tests. For pairwise comparison of the models, we refer to Appendix \ref{apd:stat_significance}.

For the unsupervised task of patient subtyping, which involves partitioning the data into $k$ distinct clusters, the quality of the resulting clusters was quantitatively measured using Silhouette scores. We employed three prominent clustering methods for assigning cluster labels to the samples: $K$-means, Spectral clustering, and Gaussian Mixture Models (GMM). The determination of the optimal number of clusters, $k$, was achieved by identifying the minimum value derived from the Elbow method, a heuristic approach commonly used to find a suitable number of clusters. 

Furthermore, in a separate case study utilizing the pediatric dataset, we leveraged the predefined obesity categories as ground truth assignments. Patients were grouped in \textit{Class I Obesity}: BMI at or above the $95$th percentile, but less than $120\%$ of the $95$th percentile, \textit{Class II Obesity}: BMI at or above $120\%$ but less than $140\%$ of the $95$th percentile and \textit{Class III Obesity}: A BMI at or above $140\%$ of the $95$th percentile \citep{hampl2023clinical}. The agreement between our derived clustering assignments and the true assignments was then analyzed using the Adjusted Rand Index (ARI), with a value of $+1$, indicating complete agreement between the predicted cluster label and the ground truth.

\subsection{Implementation Details} 
We trained the models using a fixed batch size of $32$ with the Adam optimizer and training stopped after no improvement in evaluation metrics for 10 epochs with hyperparameter tuning, hyperparameter values provided in Appendix \ref{apd:hyperparameters}. All the implementations are adapted from the official codes of the models to PyTorch modules. Preprocessing and overall customizations were adapted from a publicly available code base \citep{tipirneni2022self}. For subtyping methods, we repurposed the notebooks from \url{https://github.com/helgeingvart/phenotypeTrajectories}. The experiments were conducted on a single Tesla T4 tensor core GPU on the Amazon Web Services (AWS) Sagemaker platform.  We present the training time for both transformer-based and sub-quadratic models using the Pediatric dataset (the dataset having the longest sequence) in Appendix \ref{apd:train_time}.

\section{Results} 

\subsection{Subtyping Evaluation} 
To assess the efficacy of the learned representations for patient subtyping, we conducted a comprehensive evaluation of clustering quality across various methods and datasets. The primary internal validation metric employed was the Silhouette score, which quantifies the compactness and separation of clusters. 

\renewcommand{\arraystretch}{1.0}
\begin{table}[htbp]
  \centering 
  \caption{Silhouette scores for different clustering methods [SC: Spectral Clustering, GMM: Gaussian Mixture Model]}
\begin{tabular}{l|c|c|c}
  \toprule
     Model& $K$-Means& SC & GMM\\
    \midrule
\multicolumn{4}{c}{PhysioNet 2012 : mortality prediction} \\
\hline
\texttt{STraTS} & $.367 \pm.06$ & $.175 \pm.03$ & $.137 \pm.08$\\ 
\texttt{EHR-Mamba} &$.323 \pm.02$ & $.273 \pm.06$ & $.295 \pm.01$\\
\texttt{DuETT}& $.361 \pm.02$&$.234 \pm.03$& $.195 \pm.01$\\
\textbf{Triplet-Mamba}& $\textbf{.402} \pm\textbf{.02}$&$\textbf{.342} \pm\textbf{.01}$& $\textbf{.586} \pm\textbf{.09}$\\
  \hline
\multicolumn{4}{c}{MIMIC-IV : mortality prediction} \\
\hline
\texttt{STraTS} & $.227 \pm.03$&$.093 \pm.05$& $.129 \pm.02$\\ 
\texttt{EHR-Mamba}& $.239 \pm.09$&$.155 \pm.03$& $.212 \pm.03$\\
\texttt{DuETT}& $.331 \pm.09$&$.134 \pm.04$& $.239 \pm.06$\\ 
\textbf{Triplet-Mamba}&$\textbf{.363} \pm\textbf{.03}$&$\textbf{.274} \pm\textbf{.01}$& $\textbf{.343} \pm\textbf{.03}$\\
  \hline
\multicolumn{4}{c}{Pediatric : weight-loss prediction} \\
\hline
  \texttt{STraTS} & $.468 \pm.05$&$.283 \pm.07$& $.541 \pm.05$\\
\texttt{EHR-Mamba} & $.378 \pm.09$ & $.234 \pm.06$ & $.289 \pm.10$\\ 
\texttt{DuETT} & $.471 \pm.09$&$.261 \pm.10$& $.543 \pm.03$\\
\textbf{Triplet-Mamba}& $\textbf{.499} \pm\textbf{.08}$&$\textbf{.274} \pm\textbf{.06}$& $\textbf{.683} \pm\textbf{.05}$\\
  \bottomrule
  \end{tabular}
  \label{tab:table2} 
\end{table}

\renewcommand{\arraystretch}{1.2}
\begin{table*}[htbp]
\centering
\caption{Performance comparison against the baselines.  Mean $\pm$ STD. Boldface is best, and second-best is underlined.$^{*}$ denotes $p < 0.05$, $^{**}$ denotes $p < 0.005$ compared to \texttt{Triplet-Mamba}}.
\begin{tabular}{l|c|c|c|c}
\toprule
Metric & \texttt{STraTS} & \texttt{EHR-Mamba}  & \texttt{DuETT}  & \textbf{Triplet-Mamba} \\
\midrule
\multicolumn{5}{c}{PhysioNet 2012 : mortality prediction} \\
\hline
AUROC & $.838$ \(\pm\) $.01^*$ & .844 \(\pm\) .01 & \textbf{.857 \(\pm\) .02} & \underline{.851 \(\pm\) .02} \\
AUPRC & .487 \(\pm\) .01 & .534 \(\pm\) .03 & \textbf{.598 \(\pm\) .06} & \underline{.590 \(\pm\) .17} \\
\hline
\multicolumn{5}{c}{MIMIC-IV : mortality prediction} \\
\hline
AUROC & $.842$ \(\pm\) $.01^{**}$ & \underline{$.881$ \(\pm\) $.02^{**}$} & $.852$ \(\pm\) $.01^{**}$ & \textbf{.896 \(\pm\) .01} \\
AUPRC & .595 \(\pm\) .02 & \underline{.637 \(\pm\) .09} & .605 \(\pm\) .05 & \textbf{.639 \(\pm\) .04} \\
\hline
\multicolumn{5}{c}{Pediatric : weight-loss prediction} \\
\hline
AUROC & $.692$ \(\pm\) $.01^{**}$ & \underline{$.704$ \(\pm\) $.01^{**}$} & $.654$ \(\pm\) $.01^{**}$ & \textbf{.724 \(\pm\) .01} \\
AUPRC & \underline{.295 \(\pm\) .01} & .284 \(\pm\) .01 & .257 \(\pm\) .05 & \textbf{.301 \(\pm\) .01} \\
\bottomrule
\end{tabular}
\label{tab:table1}
\end{table*}

Table \ref{tab:table2} presents the Silhouette scores on clustering from the embeddings learned from the top-$4$ predictive performing models: STraTS, EHR-Mamba, DuETT, and our method, \texttt{Triplet-Mamba}, across the PhysioNet 2012, MIMIC-IV, and Pediatric Weight datasets. Clusters generated from the representations learned by our proposed \texttt{Triplet-Mamba} consistently achieved higher Silhouette scores, indicating its superior ability to generate embedded representations that yield more coherent and well-separated patient subtypes.

Beyond these internal validation metrics, the quality of clustering for this dataset was further corroborated using ARI, and for the pediatric weight management case study, the predefined obesity categories served as ground truth assignments, allowing the ARI to provide an objective measure of how well the learned clusters align with established clinical classifications. Table \ref{tab:table4} provides a comparison of the ARI scores, where it is evident that clustering on the representations from our model \texttt{Triplet-Mamba} has a higher score respectively for $K$-means, Spectral clustering, and GMM. The consistent outperformance of \texttt{Triplet-Mamba} across both internal (Silhouette) and external (ARI for pediatric dataset) validation metrics highlights its effectiveness in learning meaningful and clinically relevant patient subtypes from complex EHR data.

\begin{table}[htbp]
    \centering
    \caption{Adjusted Rand Index (ARI) of different clustering methods using learned representations [SC: Spectral Clustering, GMM: Gaussian Mixture Model]}
    \label{tab:table4}
    \begin{tabular}{l|c|c|c}
    \toprule
    \textbf{Model} & \textbf{K-means} & \textbf{SC} & \textbf{GMM} \\
    \midrule
    \texttt{DuETT}             & $0.632 \pm 0.02$ & $0.612 \pm 0.04$ & $0.701 \pm 0.03$ \\
    \hline
    \texttt{EHR-Mamba}         & $0.621 \pm 0.03$ & $0.611 \pm 0.02$ & $0.691 \pm 0.02$ \\
    \hline
    \textbf{Triplet-Mamba}     & $\textbf{0.683} \pm \textbf{0.05}$ & $\textbf{0.631} \pm \textbf{0.02}$ & $\textbf{0.722} \pm \textbf{0.05}$ \\
    \bottomrule
    \end{tabular}
\end{table}

To evaluate the reliability and robustness of the discovered patient subtypes, we performed a bootstrap stability analysis ($B=100$) on the Pediatric Weight Management dataset. This procedure involved re-training the Triplet-Mamba model on resampled versions of the cohort to ensure the learned manifold and resulting partitions were not artifacts of specific data subsets. We quantified the stability of the cluster assignments by calculating the mean Jaccard Similarity and the bootstrap-based ARI. Details on the stability test are given in Appendix \ref{apd:stability}.

We analyzed the clusters generated by the Gaussian Mixture Model (GMM) method to discern their characteristics. Table \ref{tab:table5} below provides a condensed version of the analysis, detailing the average maximum percentage of weight loss as well as the average duration of anti-obesity drug administration. The patients in cluster 3 exhibited the lowest average duration of treatment and the least weight loss. Further investigation revealed that the majority of patients in cluster 3 were prescribed Phentermine and similar drugs, while those in cluster 4 received GLP-1 type medications and achieved the highest weight loss.

\begin{table}[htbp]
    \centering
    \caption{Analysis of clusters from Pediatric dataset}
    \begin{tabular}{l|c|c|c|c}
    \toprule
    \textbf{Mean} & \textbf{Cluster 1} & \textbf{Cluster 2} & \textbf{Cluster 3} & \textbf{Cluster 4}\\
    \midrule
    \% of weight loss   & $4.68$ & $4.92$ & $3.94$ & $5.74$ \\
    Duration (days)     & $195.54$ & $237.48$ & $172.18$ & $204.91$ \\
    \bottomrule
    \end{tabular}
    \label{tab:table5}
\end{table}

\subsection{Predictive Performance Evaluation}
We present the prediction performance of baseline models and our model, averaged over the 10 runs in Table \ref{tab:table1}. Our model, \texttt{Triplet-Mamba}, achieves the best performance on MIMIC-IV and pediatric dataset tasks while coming close second place in the PhysioNet-2012 challenge task. Specifically, in the case of mortality prediction in the PhysioNet-2012, our model falls short of the best result of DuETT \citep{labach2023duett} by $0.006$ in AUROC. However, our model boasts a $1.7\%$ increase in AUROC and a $0.31\%$ increase in AUPRC compared to the next best results for the task of mortality prediction on the  MIMIC-IV dataset. 

Our model also outperforms other baseline models on the task of predicting maximum weight loss of $5\%$ in the pediatric weight loss dataset. It achieved a $2.8\%$ increase in AUROC and a $2.03\%$ increase in AUPRC. EHR-Mamba \citep{fallahpour2024ehrmamba} performed consistently in all the cases, achieving competitive results. However, for longer sequenced pediatric datasets as well as in the PhysionNet-2012 dataset, transformer-based STraTS performed better than the other transformer-based model DuETT \citep{labach2023duett}.

\subsection{Ablation Study} 
To ascertain the individual contributions of key components within our proposed \texttt{Triplet-Mamba} model, an ablation study was conducted. This involved systematically removing specific architectural elements from the full model and evaluating the resultant performance (AUROC) on the binary predictive tasks. Three distinct ablation scenarios were investigated: removal of the triplet embedding module, exclusion of the fusion attention layer, and absence of self-supervised pretraining.

\begin{table}[htbp]
    \centering
    \caption{Performance analysis with ablations (AUROC).}
    \small
    \begin{tabular}{l|c|c|c}
        \toprule
        \textbf{Ablations} & \textbf{PhysioNet 2012} & \textbf{MIMIC-IV} & \textbf{Pediatric Weight} \\
        \midrule

        w/o Triplet embedding & .749 \(\pm\) .02 & .802 \(\pm\) .09 & .654 \(\pm\) .03 \\
        \hline
        w/o Attention fusion & .815 \(\pm\) .06 & .858 \(\pm\) .09 & .638 \(\pm\) .02 \\
        \hline
        w/o Self-supervised Pretraining & .801 \(\pm\) .01 & .826 \(\pm\) .05 & .577 \(\pm\) .05 \\
        \hline
        \textbf{Triplet-Mamba} & \textbf{.851 \(\pm\) .02} & \textbf{.896 \(\pm\) .01} & \textbf{.724 \(\pm\) .01} \\
        \bottomrule
    \end{tabular}
    \label{tab:table3}
\end{table}

The results, summarized in Table \ref{tab:table3}, indicate that both the triplet embedding and the self-supervised pretraining greatly contribute to the model's overall performance. Their removal consistently led to a noticeable decline in AUROC across all datasets, highlighting their importance in learning robust and effective representations. Crucially, the fusion-attention layer plays a decisive role in maximizing this potential by effectively synthesizing the learned contextual embeddings; its removal resulted in a consistent decline in AUROC across all datasets, confirming its necessity in aggregating temporal dependencies for the final prediction. These findings underscore the critical role of triplet embeddings and self-supervised pretraining in the superior performance achieved by our model.

\section{Discussion} 
The key contribution of this work is the utility of the learned embeddings for patient subtyping. The subtyping evaluation in Table \ref{tab:table2} demonstrates that \texttt{Triplet-Mamba} produces a more structured latent space than baseline architectures. The higher Silhouette scores and strong ARI on external pediatric datasets indicate that our triplet embedding and self-supervised training objective effectively regularize the manifold, ensuring that clinically similar patients are mapped to proximal coordinates. These findings suggest that \texttt{Triplet-Mamba} transcends superficial label memorization, instead encoding a structured latent representation that effectively captures the longitudinal dynamics of clinical disease progression.

The results of subtype stability tests demonstrate high partition consistency across resampling iterations. \texttt{Triplet-Mamba} achieved a mean Jaccard Index of $0.814 \pm 0.03$, significantly exceeding the standard $0.75$ threshold typically indicative of stable clinical phenotyping. Furthermore, the ARI between the reference clustering and bootstrap runs remained high and consistent ($0.738\pm0.04$). This stability is particularly notable given that the subtyping was performed using only lab readings and vital signs, excluding the weight measurements used for ground-truth definitions. Our model demonstrated significantly higher stability than baselines on the MIMIC-IV and Pediatric cohorts (global $p < 0.001$). Post-hoc tests confirmed this statistical superiority ($p < 0.01$), indicating that the Mamba backbone provides a more robust latent space for longitudinal subtyping than traditional attention-based architectures.

Our experimental results also validate the efficacy of \texttt{Triplet-Mamba} as a robust framework for long-sequence clinical modeling. As evidenced in Table \ref{tab:table1}, \texttt{Triplet-Mamba} achieves competitive performance across diverse clinical tasks. In comparison, the marginal AUROC gap in the PhysioNet-2012 mortality task suggests that traditional attention mechanisms still perform well on shorter, high-density benchmarks. \texttt{Triplet-Mamba}’s superior generalizability across irregular datasets highlights its resilience to noise and data sparsity.

The model’s ability to capture intricate temporal dependencies suggests that the State Space Model (SSM) backbone effectively acts as a continuous-time filter, which is particularly suited for the non-uniform sampling rates inherent in EHR data. We recognize, however, that situating these findings within the broader landscape of high-performing traditional methods will strengthen the work, and we present the performance comparison of our model against traditional methods on PhysioNet-2012 and MIMIC-IV datasets in Appendix \ref{apd:baseline_comp}.

The ablation study (Table \ref{tab:table3}) confirms that the synergy between triplet embeddings and self-supervised pretraining is the primary driver of performance. We observe that: 1) Triplet Embeddings provide the necessary contrastive signal to distinguish between subtle clinical state transitions. 2) Self-supervised Pretraining allows the model to build a robust prior from unlabeled sequences, addressing the relatively smaller data problem common in specialized clinical domains. 3)  Unlike Transformer-based models that suffer from $O(L^2)$ complexity, the underlying SSM in \texttt{Triplet-Mamba} scales linearly ($O(L)$) with sequence length. This computational efficiency is not just an optimization but a functional requirement for processing long-term patient histories that could otherwise exceed the memory constraints of standard attention blocks.

\paragraph{Limitations}

Despite the performance gains, certain challenges remain. The fusion attention layer, while theoretically sound for multi-modal integration, showed a moderate contribution in our current ablation, suggesting that more complex fusion strategies (e.g., cross-mamba blocks) could be explored. Furthermore, while \texttt{Triplet-Mamba} captures long-range dependencies, the interpretability of its hidden states—unlike the direct visualization of attention maps—remains an open research question. Our model embeds EHR events in observation triplets while combining static variables, such as demographics, separately with the temporal encoding through concatenation. Due to the unavailability of special types of events, such as procedures and diagnoses, in the PhysioNet-2012 dataset, our model only encoded the measurement values from the EHR. For the pediatric weight management dataset, we only considered lab readings and measurement values for predicting weight loss. However, the original dataset contains records of procedures, diagnoses, medications, etc., which we expect to incorporate in our future work. With more available information about patient records, our long-context sub-quadratic model will presumably understand patient condition progression and identify latent subgroups among them. In its present form, the proposed model does not include text or imaging data. Nonetheless, integrating multi-modal information into this model would further enhance its generalizability and potentially its performance.

\bibliographystyle{ACM-Reference-Format}
\bibliography{bibliography}

\newpage
\appendix
\section{Mamba block and SSM unit}\label{apd:mamba_block}
\begin{figure}[htbp]
    \centering 
    \includegraphics[width=0.5\textwidth]{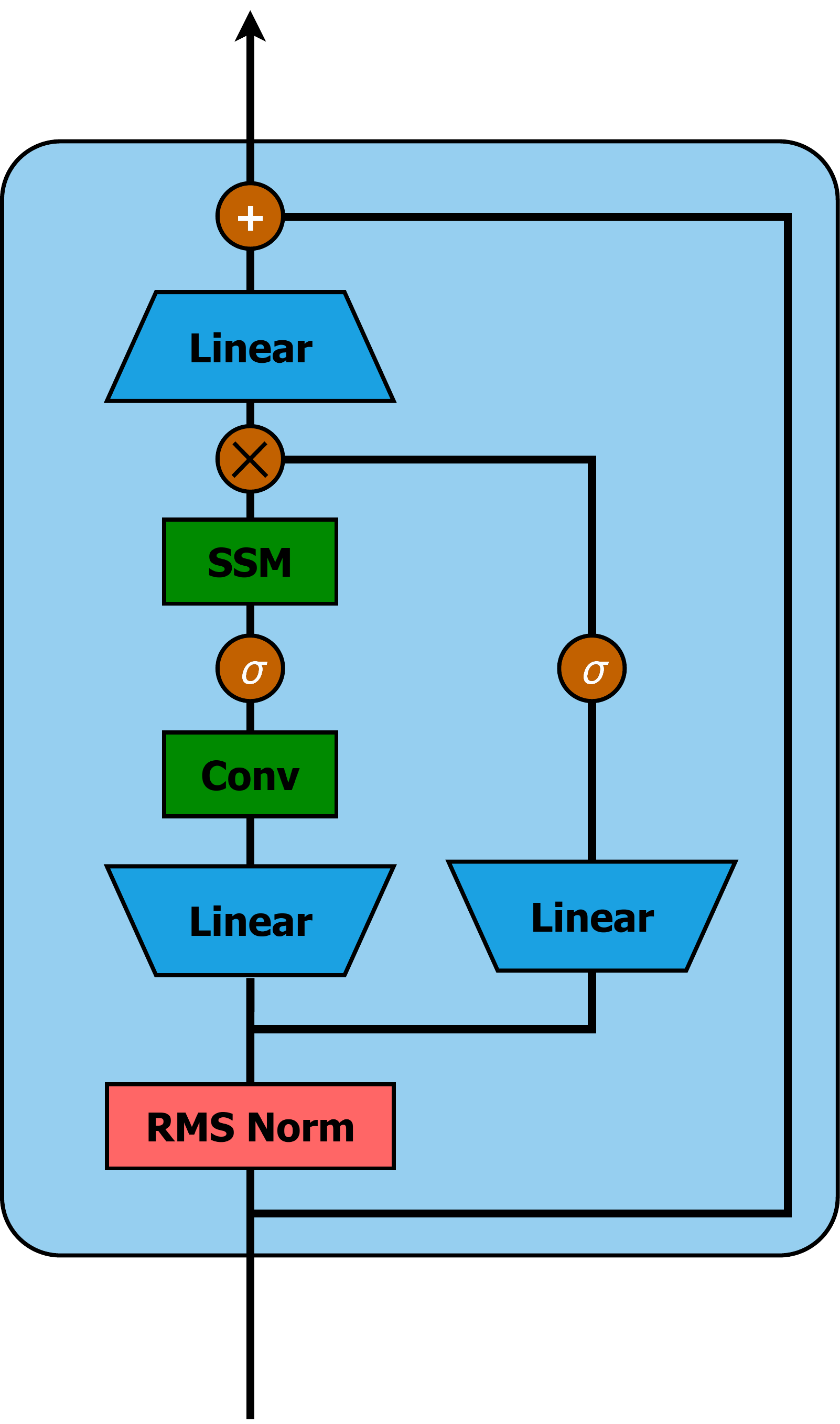}
    \caption{Mamba block}
    \label{fig:mamba} 
\end{figure}

\paragraph{SSM unit} The Mamba block (shown at the right section of figure \ref{fig:mamba} consists of a SSM unit which maps an input sequence $x(t)\in \mathbb{R}$ to an output sequence $y(t)\in \mathbb{R}$ using an implicit hidden state $h(t)\in \mathbb{R}^\psi$ with $\psi$ being the state size. The mapping is defined by the following linear differential equations: 
\begin{equation}
h'(t)=\textbf{A}h(t)+\textbf{B}x(t)
\end{equation}
\begin{equation}
    y(t)=\textbf{C}h(t)
\end{equation}Here, $\textbf{A}\in {\mathbb{R}}^{\psi\times \psi}, \textbf{B}\in {\mathbb{R}}^{\psi \times 1},$ and $\textbf{C}\in {\mathbb{R}}^{1\times \psi}$ are learnable matrices. For multidimensional sequences, this system is applied independently to each dimension \citep{guefficiently}. To apply SSMs to discrete sequences, the system is discretized using a step size $\Delta$:
\begin{equation}
    h_t=\overline{\textbf{A}}h_{t-1}+\overline{\textbf{B}}x_t
\end{equation}
\begin{equation}
    y_t=\textbf{C}h_t
\end{equation}The discrete parameters $(\overline{\textbf{A}}, \overline{\textbf{B}})$ are derived using the zero-order hold (ZOH) rule \citep{guefficiently, gumamba}:
\begin{equation}
    \overline{\textbf{A}}=\exp{(\Delta \textbf{A})}
\end{equation}
\begin{equation}
    \overline{\textbf{B}}={(\Delta \textbf{A})}^{-1}((\exp{(\Delta \textbf{A})-\textbf{I})\cdot \Delta \textbf{B} }
\end{equation}
The model uses a global convolution kernel $\overline{K}$ to enable parallel sequence processing and scaling training following the equation \citep{gumamba}:
\begin{equation}
    \overline{K}=(C\overline{\textbf{B}}, C\overline{\textbf{AB}}, \dots, C\overline{\textbf{A}}^k\overline{\textbf{B}},\dots) 
\end{equation}
\begin{equation}
    y=x *\overline{K}
\end{equation}

\section{Training Time}\label{apd:train_time}
We present the training time for both transformer-based and sub-quadratic models using the Pediatric dataset (the dataset having the longest sequence):

\begin{table}[htbp]
    \centering
    \caption{Model training time comparison}
    \begin{tabular}{l|c}
    \toprule
    \textbf{Model} & \textbf{Average Training Time (minutes)} \\
    \midrule
    \texttt{STraTS}             & $26.17 \pm 1.2$ \\
    \hline
    \texttt{DuETT}             & $30.83 \pm 2.5$ \\
    \hline
    \texttt{EHR-Mamba}         & $13.46 \pm 1.4$ \\
    \hline
    \textbf{Triplet-Mamba}     & $5.25 \pm 2.8$  \\
    \bottomrule
    \end{tabular}
    \label{tab:appC}
\end{table}

\section{Extended cohort description}\label{apd:extended_cohort}
\begin{itemize}
    \item PhysioNet challenge is a publicly available standardized dataset with the task of predicting in-hospital mortality after the first $48$ hours of patient stays in the ICU, with $14.2\%$ of positive labels. The dataset consists of $11,988$ patients with $42$ different variables, including $37$ time series event-types. We process the data according to the previous work \citep{tipirneni2022self} for predicting mortality. The set of observation windows is defined (in hours) as the first 48 hours of the patient's stay in ICU for forecasting windows and the prediction window is the $2$-hour period following the observation window. The data from set-b and set-c together is split into training and validation $(80:20)$ while set-a is used for testing purposes.
    \item MIMIC-IV is a public dataset containing retrospective, de-identified data patients at the Beth Israel Deaconess Medical Center. We evaluated tasks on a derived ICU dataset, containing $53,150$ patients with $69,211$ admissions. We followed prior work \citep{labach2023duett} in defining mortality prediction in the ICU dataset, where we use the first 48 hours of the patient stay as the input time window, predicting whether death occurs later during the hospital stay. We use a patient-level $60\%:20\%:20\%$ split between the training, validation, and test sets.
    \item Pediatric weight management dataset was extracted from the EHR of Nemours Children's Health, which is a large pediatric healthcare system providing primary, specialty, inpatient, and emergency care to pediatric patients across five US states. The EHR data was available as a de-identified dataset using the pediatric-specific Common Data Model (CDM) is anchored by patients and their encounters with the health system and is based on the Observational Medical Outcomes Partnership (OMOP) \citep{makadia2014transforming} model. The target task is designed to predict $5\%$ maximum weight loss using anti-obesity medications. With $18.7\%$ positive labels and a maximum time series of $8,654$ events and $7,183$ days, this dataset contains $14,392$ individual pediatric records \citep{dip2026pedobesity, pmlr-v259-fayyaz25a}.
\end{itemize}

\section{Extended Baseline Models}\label{apd:baseline_model_comp}
Here is a complete list of all $9$ baseline models used for evaluation with results in Table \ref{tab:appA}.

    \paragraph{Gated Recurrent Unit (GRU)} \citep{chung2014empirical} Hourly aggregated variables are packed into a time-series matrix for input and the final hidden state is transformed by a dense layer to generate output.
    \paragraph{Temporal Convolutional Network (TCN)} \citep{bai2018empirical} Input similar to GRU which is passed through a stack of temporal convolution layers with residual connections. The representations from the last time step of the last layer is transformed by a dense layer to generate output. 
    \paragraph{Simply Attend and Diagnose (SaND)} \citep{song2018attend} This model utilizes an input representation identical to that of the GRU. The input is processed through a Transformer that employs causal attention, followed by a dense interpolation layer.
    \paragraph{GRU with trainable Decays (GRU-D)} \citep{che2018recurrent} The GRU-D cell processes a vector of variable values at each time step, modifying the standard GRU cell. It decays unobserved values to global means and adjusts the hidden state based on the time since the last observation of each variable.
    \paragraph{Interpolation-prediction Network (InterpNet)} \citep{shuklainterpolation} This model features a semi-parametric interpolation network that estimates all variables at regular, predefined time intervals, followed by a prediction network based on a GRU. Additionally, it employs a reconstruction loss to enhance the performance of the interpolation network.
    \paragraph{Set Functions for Time Series (SeFT)} \citep{horn2020set} This model is designed to process a set of observation triplets. It incorporates sinusoidal encoding to effectively embed temporal information. Additionally, the deep network employed to integrate the observation embeddings is structured as a set function, utilizing a simplified yet efficient variation of multi-head attention.
    \paragraph{Self-supervised Transformer for Time-Series (STraTS)} \citep{tipirneni2022self} The model uses a continuous value embedding technique to encode continuous time and variable values without discretization. It features a Transformer component with multi-head attention layers, allowing it to learn contextual triplet embeddings while avoiding issues related to recurrence and vanishing gradients found in recurrent architectures.
    \paragraph{EHR-Mamba} \citep{fallahpour2024ehrmamba}  This model encodes EHR data in a combination of seven different embeddings and uses stacked Mamba blocks for mapping input sequence to output tensor for downstream forecasting or predicting tasks.
    \paragraph{DuETT} \citep{labach2023duett} This architecture extends transformers to exploit both time and event modalities of EHR data. Each DuETT layer is made up of two Transformer sub-layers that attend along the event and time dimensions, respectively.  The representation learned from the time and event embeddings is used for downstream tasks by attaching respective task heads.

\section{Performance comparison against traditional methods.}\label{apd:baseline_comp}
The predictive performance of the models, including all nine baseline models, is shown in Table \ref{tab:appE}.

\renewcommand{\arraystretch}{1.2}
\begin{table*}[htbp]
  \centering 
  \caption{Performance comparison of \texttt{Triplet-Mamba} against traditional methods with cross-validation. Best results are highlighted with bold face.}
  \begin{tabular}{p{2.0cm}|l|c}
  \toprule
     \textbf{Dataset} & \textbf{Model} & \textbf{AUROC} \\
    \midrule
    \multirow{7}{10em}{PhysioNet} & Logistic Regression & $0.838 \pm 0.017$ \\ 
        & Gaussian Na\"{\i}ve Bayes & $0.783 \pm 0.025$ \\
        & SVM                       & $0.817 \pm 0.054$ \\ 
        & Decision Tree             & $0.793 \pm 0.068$ \\
        & Random Forest             & $0.848 \pm 0.032$ \\
        & AdaBoost                  & $0.783 \pm 0.023$ \\
        & \texttt{Triplet-Mamba} (ours)   & \textbf{$0.851$} $\pm$ \textbf{$0.021$} \\
    \hline
    \multirow{7}{10em}{MIMIC-IV} & Logistic Regression & $0.844 \pm 0.023$ \\ 
        & Gaussian Na\"{\i}ve Bayes & $0.834 \pm 0.024$ \\
        & SVM                       & $0.860 \pm 0.054$ \\ 
        & Decision Tree             & $0.852 \pm 0.018$ \\
        & Random Forest             & $0.861 \pm 0.038$ \\
        & AdaBoost                  & $0.878 \pm 0.036$ \\
        & \texttt{Triplet-Mamba} (ours)    & $0.896 \pm 0.013$ \\
  \bottomrule
  \end{tabular}
  \label{tab:appE} 
\end{table*}

\section{Statistical Evaluation}\label{apd:stat_significance}
We conducted independent experiments to evaluate the significance of the results. Independent two-sample t-test results for pairwise comparison of our model against other baseline results are shown in Table \ref{tab:appF}.

\begin{table*}[htbp]
\centering
\caption{Statistical test results comparing \texttt{Triplet-Mamba} against baseline models. $^{*}$ denotes that the individual comparison is statistically significant.}
\resizebox{\columnwidth}{!}{
\begin{tabular}{l|c|c|c|c}
\toprule
\textbf{Model comparison} & \textbf{Group 1 mean AUROC} & \textbf{Group 1 mean AUROC}  & \textbf{t-statistics}  & \textbf{p-value} \\
\midrule
\multicolumn{5}{c}{\textbf{PhysioNet 2012}} \\
\hline
\texttt{Triplet-Mamba} vs \texttt{STraTS} & $0.851$ & $0.838$ & $2.5465$ & $0.01012^*$ \\
\texttt{Triplet-Mamba} vs \texttt{EHR-Mamba} & $0.851$ & $0.844$ & $1.0203$ & $0.16056$ \\
\texttt{Triplet-Mamba} vs \texttt{DueTT} & $0.851$ & $0.857$ & $0.1651$ & $0.43535$ \\
\hline
\multicolumn{5}{c}{\textbf{MIMIC-IV}} \\
\hline
\texttt{Triplet-Mamba} vs \texttt{STraTS} & $0.896$ & $0.842$ & $12.5150$ & $<0.00001^*$ \\
\texttt{Triplet-Mamba} vs \texttt{EHR-Mamba} & $0.896$ & $0.881$ & $3.3883$ & $0.00164^*$ \\
\texttt{Triplet-Mamba} vs \texttt{DueTT} & $0.896$ & $0.852$ & $9.9323$ & $<0.00001^*$ \\
\hline
\multicolumn{5}{c}{\textbf{Pediatric Dataset}} \\
\hline
\texttt{Triplet-Mamba} vs \texttt{STraTS} & $0.724$ & $0.692$ & $5.3954$ & $0.00002^*$ \\
\texttt{Triplet-Mamba} vs \texttt{EHR-Mamba} & $0.724$ & $0.704$ & $3.9952$ & $0.00042^*$ \\
\texttt{Triplet-Mamba} vs \texttt{DueTT} & $0.724$ & $0.654$ & $14.8695$ & $<0.00001^*$ \\
\bottomrule
\end{tabular}
}
\label{tab:appF}
\end{table*}

\renewcommand{\arraystretch}{1.2}
\section{Subtype stability analysis.}\label{apd:stability}
The one-way Analysis of Variance (ANOVA) table summarizes the statistical evaluation of clustering stability across 100 bootstrap resampling iterations. In this analysis, the model architecture serves as the categorical independent variable, while the ARI constitutes the dependent measure of representation reliability. The global $p$-value assesses the null hypothesis that all models exhibit equivalent stability under data perturbation. Significant results ($p < 0.05$) indicate that architectural design choices fundamentally impact the consistency of discovered phenotypes, while subsequent pairwise post-hoc tests provide a rigorous mathematical basis for identifying the statistical superiority of specific models in capturing robust clinical trajectories, shown in Table \ref{tab:appStability}.

\begin{table}[htbp]
\centering
\caption{One-way ANOVA and Post-hoc Pairwise Comparisons of Clustering Stability (ARI) across 100 Bootstrap Iterations.}
\label{tab:anova_stability_results}
\begin{tabular}{@{}l|c|c|c|c@{}}
\toprule
Dataset & Global ANOVA ($p$-value) & vs. STraTS & vs. EHR-Mamba & vs. DuETT \\ \midrule
PhysioNet 2012 & $p=0.122$ & $p>0.05$ & $p>0.05$ & $p>0.05$ \\
MIMIC-IV & $p<0.001$ & $p<0.01$ & $p=0.198$ & $p<0.01$ \\
Pediatric & $p<0.001$ & $p<0.01$ & $p<0.01$ & $p<0.01$ \\ \bottomrule
\end{tabular}
\label{tab:appStability}
\end{table}

\section{Cluster evaluations}
Below are the distribution and balance statistics for the discovered GMM clusters across all evaluation cohorts.

\begin{table}[htbp]
\centering
\caption{Cluster distribution and dominant characteristics}
\label{tab:cluster_analysis}
\resizebox{\columnwidth}{!}{ 
\begin{tabular}{@{}l|c|l@{}}
\toprule
\textbf{Cluster ID} & \textbf{Count (\%)} & \textbf{Dominant characteristics} \\ \midrule
\multicolumn{3}{c}{ \textbf{PhysioNet (N=$11,988$)}} \\ \hline
Cluster 1 & $5,215$ $(43.5\%)$ & Low variance in lab measurements, baseline risk tier \\
Cluster 2 & $4,615$ $(38.5\%)$ & Elevated BUN, serum creatinine and fluctuating potassium trajectories \\
Cluster 3 & $2,158$ $(18.0\%)$ & Highly elevated heart rates, depressed oxygen saturation, high mortality rate \\ \hline
\multicolumn{3}{c}{ \textbf{MIMIC-IV (N=$69,211$)}} \\ \hline
Cluster 1 & $31,837$ $(46.0\%)$ & Lower mortality endpoint incidence \\
Cluster 2 & $24,520$ $(35.4\%)$ & High variance in blood pressure parameters, elevated pulse, frequent cardiac unit assignment \\
Cluster 3 & $12,854$ $(18.6\%)$ & Extreme lab value missingness rates, acute physiological derangement \\ \hline
\multicolumn{3}{c}{ \textbf{Pediatric (N=$14,392$)}} \\ \hline
Cluster 1 & $4,821$ $(33.5\%)$ & Gradual BMI progression tracking closer to baseline percentiles \\
Cluster 2 & $4,246$ $(29.5\%)$ & Moderate long-term trajectory variations \\
Cluster 3 & $2,950$ $(20.5\%)$ & Extreme lab value missingness rates, acute physiological derangement \\
Cluster 4 & $2,375$ $(16.5\%)$ & Highest initial BMI parameters, longer treatment durations, maximum weight loss outcomes \\
\bottomrule
\end{tabular}
}
\end{table}

\section{Clinical validity of clusters}
To establish true clinical validity and confirm our partitions are well-distributed, we performed post-hoc statistical variance testing across several continuous and categorical clinical features. The following tables present a subset of continuous and categorical features selected based on their clinical relevance to critical care and high statistical variance (all $p<0.001$) across the groups.

\begin{table}[htbp]
\centering
\caption{Clinical Characteristics Across Clusters for Different Datasets}
\label{tab:cluster_summary}

\textbf{PhysioNet} \\[0.3em]
\resizebox{\columnwidth}{!}{
\begin{tabular}{l|c|c|c|c}
\hline
\textbf{Metric} & \textbf{Cluster 1} & \textbf{Cluster 2} & \textbf{Cluster 3} & \textbf{Test Stat} \\
\hline
Heart rate (bpm) & 81.4 $\pm$ 10.2 & 88.6 $\pm$ 12.4 & 104.2 $\pm$ 15.8 & F = 214.5 \\
$O_2$ saturation (\%) & 98.2 $\pm$ 1.1 & 96.4 $\pm$ 2.3 & 91.8 $\pm$ 4.6 & F = 189.3 \\
Serum Creatinine (mg/dL) & 0.92 $\pm$ 0.21 & 2.41 $\pm$ 0.85 & 1.65 $\pm$ 0.72 & F = 342.1 \\
In-hospital mortality (\%) & 3.2\% (167) & 15.4\% (711) & 38.3\% (826) & $\chi^2$ = 1842.6 \\
\hline
\end{tabular}
}
\vspace{1em}

\textbf{MIMIC-IV} \\[0.3em]
\resizebox{\columnwidth}{!}{
\begin{tabular}{l|c|c|c|c}
\hline
\textbf{Metric} & \textbf{Cluster 1} & \textbf{Cluster 2} & \textbf{Cluster 3} & \textbf{Test Stat} \\
\hline
Systolic BP (mmHg) & 118.5 $\pm$ 9.4 & 138.2 $\pm$ 18.6 & 94.1 $\pm$ 14.2 & F = 412.8 \\
Blood Glucose (mg/dL) & 112.4 $\pm$ 15.1 & 145.8 $\pm$ 32.4 & 188.2 $\pm$ 45.6 & F = 189.3 \\
In-hospital mortality (\%) & 4.1\% (1,305) & 12.0\% (2,940) & 29.5\% (3,688) & $\chi^2$ = 4811.2 \\
\hline
\end{tabular}
}
\vspace{1em}

\textbf{Pediatric Weight Management} \\[0.3em]
\resizebox{\columnwidth}{!}{
\begin{tabular}{l|c|c|c|c|c}
\hline
\textbf{Metric} & \textbf{C1} & \textbf{C2} & \textbf{C3} & \textbf{C4} & \textbf{Test Stat} \\
\hline
Weight Loss (\%) & 4.68 $\pm$ 1.1 & 4.92 $\pm$ 1.3 & 3.94 $\pm$ 0.9 & 5.74 $\pm$ 1.8 & F = 143.2 \\
Duration (days) & 195.5 $\pm$ 34 & 237.4 $\pm$ 41 & 172.2 $\pm$ 28 & 204.9 $\pm$ 38 & F = 98.6 \\
\hline
\end{tabular}
}
\end{table}

\section{Performance on different sequence lengths}
We present the mean AUROC on weight loss prediction in Pediatric data set stratified by sequence lengths in table \ref{tab:appD}. It is to be noted that the number of samples with 4K-8K length were minimal yet the AUROC did not degrade too much.

\begin{table}[htbp]
    \centering
    \caption{Effect of different sequence length on AUROC}
    \begin{tabular}{c|c}
    \toprule
    \textbf{Sequence length} & \textbf{Mean AUROC} \\
    \midrule
    $<500$ & $0.717 \pm 0.02$ \\
    $500-1K$ & $0.726 \pm 0.01$ \\
    $1K-2K$ & $0.738 \pm 0.02$ \\
    $2K-4K$ & $0.723 \pm 0.01$ \\
    $4K-8K$ & $0.715 \pm 0.02$ \\
    \midrule
    \textbf{Overall} & \textbf{$0.724 \pm 0.01$} \\
    \bottomrule
    \end{tabular}
    \label{tab:appD}
\end{table}

\section{Performance comparison against all models.}\label{apd:third}
The predictive performance of the models, including all nine baseline models, is shown in Table \ref{tab:appA}.
\begin{table*}[htbp]
  \centering 
  \caption{Predictive performance comparison. The results show the mean and standard deviation of the metrics with 10 repeated experiments. Best results are highlighted with bold face, second best results are underlined.}
  \begin{tabular}{l|l|c|c}
  \toprule
     \textbf{Dataset : Task} & \textbf{Model} & \textbf{AUROC} & \textbf{AUPRC}\\
    \midrule
     \multirow{10}{10em}{PhysioNet-2012 : mortality prediction} &GRU& $0.818 \pm 0.009$ &$0.407 \pm 0.007$\\ 
     &TCN& $0.801 \pm 0.002$ &$0.433 \pm 0.004$\\
  &SAnD& $0.801 \pm 0.011$&$0.404 \pm 0.019$\\ 
  &GRU-D&$0.843 \pm 0.006$ &$0.451 \pm 0.007$\\
  &InterpNet& $0.803 \pm 0.005$&$0.402 \pm 0.015$\\
  &SeFT& $0.812 \pm 0.008$&$0.464 \pm 0.012$\\
  &STraTS&$0.838 \pm 0.009$ &$0.487 \pm 0.001$\\
  &EHR-Mamba&$0.844 \pm 0.006$ &$0.534 \pm 0.026$\\
  &DuETT&$\textbf{0.857} \pm \textbf{0.018}$ &$\textbf{0.598} \pm \textbf{0.056}$\\
  &\texttt{Triplet-Mamba} (ours)&$\underline{0.851\pm 0.021}$ &$\underline{0.590 \pm 0.168}$\\
  \hline
    \multirow{10}{10em}{MIMIC-IV : mortality prediction}&GRU& $0.818 \pm 0.025$ &$0.558 \pm 0.016$\\ 
    &TCN& $0.823 \pm 0.023$ &$0.584 \pm 0.026$\\
  &SAnD& $0.816 \pm 0.016$&$0.578 \pm 0.021$\\ 
  &GRU-D&$0.824 \pm 0.045$&$0.592 \pm 0.064$\\
  &InterpNet& $0.823 \pm 0.051$&$0.587 \pm 0.021$\\
  &SeFT& $0.848 \pm 0.021$&$0.573 \pm 0.071$\\
  &STraTS&$0.842 \pm 0.003$ &$0.595 \pm 0.012$\\
  &EHR-Mamba&$\underline{0.881 \pm 0.021}$ &$\underline{0.637 \pm 0.095}$\\
  &DuETT&$0.852 \pm 0.007$ &$0.605 \pm 0.048$\\
  &\texttt{Triplet-Mamba} (ours) &$\textbf{0.896}\pm \textbf{0.004}$ &$\textbf{0.639} \pm \textbf{0.042}$\\
    \hline
    \multirow{10}{10em}{Pediatric : weight-loss prediction}&GRU& $0.653 \pm 0.002$ &$0.280 \pm 0.011$\\ 
    &TCN& $0.654 \pm 0.005$ &$0.219 \pm 0.015$\\
  &SAnD& $0.661 \pm 0.012$&$0.279 \pm 0.005$\\ 
  &GRU-D&$0.644 \pm 0.001$&$0.271 \pm 0.003$\\
  &InterpNet& $0.612 \pm 0.082$&$0.251 \pm 0.027$\\
  &SeFT& $0.644 \pm 0.007$&$0.288 \pm 0.001$\\
  &STraTS&$0.692 \pm 0.002$ &$\underline{0.295 \pm 0.002}$\\
  &EHR-Mamba&$\underline{0.704 \pm 0.006}$ &$0.284 \pm 0.012$\\
  &DuETT&$0.654 \pm 0.008$ &$0.257 \pm 0.051$\\
  &\texttt{Triplet-Mamba} (ours) &$\textbf{0.724}\pm \textbf{0.004}$ &$\textbf{0.301} \pm \textbf{0.014}$\\  
  \bottomrule
  \end{tabular}
  \label{tab:appA} 
\end{table*}

\section{Implementation details}\label{apd:hyperparameters}
Table \ref{tab:appB} lists the hyperparameters used in the experiments for all models across all the datasets.
\begin{table*}[hp]
    \centering
    \caption{Hyperparameters used for our experiments in this study.}
    \resizebox{\columnwidth}{!}{
    \begin{tabular}{l|p{4.5cm}|p{4.5cm}|p{4.5cm}}
        \toprule
        \textbf{Model} & \textbf{PhysioNet-2012} & \textbf{MIMIC-IV} & \textbf{Pediatric weight}\\
        \midrule
        GRU & units = $48$, rec d/o = $0.2$, output d/o = $0.2$, lr = $0.0001$  &units = $48$, rec d/o = $0.2$, output d/o = $0.2$, lr = $0.0001$ & units = $60$, rec d/o = $0.2$, output d/o = $0.2$, lr = $0.0001$\\\hline
        TCN & layers = $4$, filters = $64$, kernel size = $4$, d/o =$ 0.5$ &layers = $4$, filters = $64$, kernel size = $4$, d/o =$ 0.5$ &layers = $6$, filters = $128$, kernel size = $4$, d/o = $0.5$\\\hline
        SAnD & N = $4$, r = $24$, M = $12$, d/o = $0.3$, d = $64$, h = $4$   &N = $4$, r = $24$, M = $12$, d/o = $0.3$, d = $64$, h = $4$ &N = $4$, r = $24$, M = $12$, d/o = $0.3$, d = $64$, h = $4$ \\\hline
        GRU-D & units = $48$, rec d/o = $0.2$, output d/o = $0.2$, lr = $0.00001$ &units = $48$, rec d/o = $0.2$, output d/o = $0.2$, lr = $0.00001$ &units = $60$, rec d/o = $0.2$, output d/o = $0.2$, lr = $0.00001$ \\\hline
        SeFT & lr = $0.00005$, n\_phi\_layers = $4$, phi\_width = $128$, phi\_dropout = $0.2$, n\_psi\_layers = $2$, psi\_width = $64$, psi\_latent\_width = $128$, dot\_prod\_dim = $128$, n\_heads = $4$, attn\_dropout = $0.5$, latent\_width = $32$, n\_rho\_layers = $2$, rho\_width = $512$, rho\_dropout = $0.0$, max\_timescale = $100.0$, n\_positional\_dims = $4$ &lr = $0.00005$, n\_phi\_layers = $4$, phi\_width = $128$, phi\_dropout = $0.2$, n\_psi\_layers = $2$, psi\_width = $64$, psi\_latent\_width = $128$, dot\_prod\_dim = $128$, n\_heads = $4$, attn\_dropout = $0.5$, latent\_width = $32$, n\_rho\_layers = $2$, rho\_width = $512$, rho\_dropout = $0.0$, max\_timescale = $100.0$, n\_positional\_dims = $4$ &lr = $0.0001$, n\_phi\_layers = $4$, phi\_width = $128$, phi\_dropout = $0.2$, n\_psi\_layers = $2$, psi\_width = $64$, psi\_latent\_width = $128$, dot\_prod\_dim = $128$, n\_heads = $4$, attn\_dropout = $0.5$, latent\_width = $32$, n\_rho\_layers = $2$, rho\_width = $512$, rho\_dropout = $0.0$, max\_timescale = $100.0$, n\_positional\_dims = $4$\\ \hline
        InterpNet & ref\_points = $96$, units = $100$, input dim = $0.2$, rec dim = $0.2$, lr = $0.001$&ref\_points = $96$, units = $100$, input dim = $0.2$, rec dim = $0.2$, lr = $0.001$&ref\_points = $192$, units = $100$, input dim = $0.2$, rec dim =$ 0.2$, lr = $0.001$\\\hline
        STraTS& d = $32$, M = $4$, h = $8$, dropout = $0.2$, lr = $0.0005$&d = $32$, M = $4$, h = $8$, dropout = $0.2$, lr = $0.0005$&d = $32$, M= $4$, h = $8$, dropout = $0.2$, lr = $0.0005$\\\hline
        EHR-Mamba& d = $32$, M = $4$, h = $8$, dropout = $0.2$, lr = $0.0005$&d = $32$, M = $4$, h = $8$, dropout = $0.2$, lr = $0.0005$&d = $32$, M = $4$, h = $8$, dropout = $0.2$, lr = $0.0005$\\\hline
        DueTT& d = $32$, M = $4$, h = $8$, dropout = $0.2$, lr = $0.0005$&d = $32$, M = $4$, h = $8$, dropout = $0.2$, lr = $0.0005$&d = $32$, M = $4$, h = $8$, dropout = $0.2$, lr = $0.0005$\\\hline
        \texttt{Triplet-Mamba} &$\tau$ = $8$, M = $2$, h = $2$, dropout = $0.1$, lr = $0.0005$ & $\tau$ = $8$, M = $2$, h = $2$, dropout = $0.1$, lr = $0.0005$ & $\tau$ = $8$, M = $2$, h = $2$, dropout = $0.1$, lr = $0.0005$\\
    \end{tabular}
    }
    \label{tab:appB}
\end{table*}

\end{document}